\documentclass[11pt]{article}

\usepackage[utf8]{inputenc}
\usepackage{amsmath,amssymb,amsthm}
\usepackage{booktabs}
\usepackage{graphicx}
\usepackage{hyperref}
\usepackage{natbib}
\usepackage[margin=1in]{geometry}
\usepackage{algorithm}
\usepackage{algpseudocode}

\title{Massive Redundancy in Gradient Transport\\Enables Sparse Online Learning}
\author{Aur Shalev-Merin\\Independent Researcher}
\date{}

\begin{document}
\maketitle

\begin{abstract}
Real-time recurrent learning (RTRL) computes exact online gradients by propagating a Jacobian tensor forward through recurrent dynamics, but at $O(n^4)$ cost per step. Prior work has sought structured approximations (rank-1 compression, graph-based sparsity, Kronecker factorization). We show that, in the continuous error signal regime, the recurrent Jacobian is \emph{massively redundant}: propagating through a random 6\% of paths ($k{=}4$ of $n{=}64$) recovers $84 \pm 6\%$ of full RTRL's adaptation ability across five seeds, and the absolute count $k{=}4$ remains effective from $n{=}64$ to $n{=}256$ (6\% $\to$ 1.6\%, recovery $84 \to 78\%$). Sparse RTRL gets relatively cheaper as networks grow, with modest degradation in recovery. In RNNs, the recovery is selection-invariant (even adversarial path selection works) and exhibits a step-function transition from zero to any nonzero propagation. Spectral analysis reveals the mechanism: the Jacobian is full-rank but near-isotropic (condition numbers 2.6--6.5), so any random subset provides a directionally representative gradient estimate. On chaotic dynamics (Lorenz attractor), sparse propagation is more numerically stable than full RTRL (CV 13\% vs.\ 88\%); subsampling avoids amplifying pathological spectral modes. On non-chaotic tasks, full RTRL is stable. The redundancy extends to LSTMs ($k{=}4$ matches full RTRL) and to transformers via sparse gradient transport (50\% head sparsity outperforms the dense reference; 33\% is borderline). The higher transformer threshold reflects head specialization, not isotropy. On real primate neural data, sparse RTRL ($k{=}4$) adapts online to cross-session electrode drift ($80 \pm 11\%$ recovery, 5 seeds), where sparse propagation is again more stable than full RTRL. Without continuous error signal, Jacobian propagation accumulates numerical drift and degrades all RTRL variants, a scope condition for all forward-mode methods. Results hold with SGD ($92 \pm 1\%$ recovery on one task) and appear independent of optimizer choice.
\end{abstract}

\section{Introduction}
\label{sec:introduction}

Online learning in recurrent networks requires solving a credit assignment problem: how does changing a weight now affect outputs in the future? Two classical approaches bracket the design space. Backpropagation through time \citep[BPTT;][]{werbos1990bptt} stores a trajectory of hidden states and backpropagates through the unrolled computation graph. This is effective but fundamentally offline and memory-intensive. Real-time recurrent learning \citep[RTRL;][]{williams1989rtrl} propagates a Jacobian tensor forward in time, tracking how each parameter affects the current hidden state through the entire history of recurrent dynamics. RTRL is truly online (no stored trajectories, no discrete training phases), but its $O(n^4)$ per-step cost has limited it to small networks.

Between these extremes lies a spectrum of approximations. At the cheap end, eligibility traces \citep{murray2019rflo, bellec2020eprop} drop the Jacobian propagation term entirely, reducing cost to $O(n^2)$ but losing the ability to assign credit through recurrent connections. At the expensive end, exact methods like \citet{benzing2019ok} achieve $O(n^3)$ with Kronecker structure. In between, UORO \citep{tallec2017uoro} compresses the Jacobian to rank 1, SnAp \citep{menick2021snap} exploits graph structure in weight-sparse networks, and element-wise architectures \citep{irie2024elstm} eliminate inter-neuron recurrence to make RTRL tractable. Each method proposes a specific structured approximation with specific tradeoffs between cost, bias, and variance.

We ask a more basic question: \textbf{how much of the recurrent Jacobian is actually needed?} Rather than designing a new approximation scheme, we test the simplest possible one: randomly dropping most of the Jacobian propagation paths and keeping a small arbitrary subset. If this works, the information in the Jacobian is redundant, and the structured design choices in prior work may be solving a problem that does not need solving.

We find massive redundancy, in the regime where RTRL is effective (continuous, dense error signals). On dynamical prediction tasks with online distribution shifts, propagating through a random 4 out of 64 Jacobian paths (6\%) recovers 78--84\% of full RTRL's adaptation ability (log-scale recovery across 5 seeds). The recovery is robust across tasks (sine waves, chaotic Lorenz attractors, multi-regime dynamics) and architectures (vanilla RNNs and LSTMs with gate Jacobian sparsification). It follows an absolute-$k$ scaling law: $k{=}4$ works at $n{=}64$ (6\%), $n{=}128$ (3\%), and $n{=}256$ (1.6\%), with recovery remaining substantial ($78$--$84\%$) despite the fraction decreasing $4\times$. The required number of paths does not grow with network size, so sparse RTRL gets relatively cheaper at larger scales. The recovery is also selection-invariant: deliberately choosing the \emph{weakest} recurrent connections works as well as choosing the strongest. Even adversarial path selection cannot break the approximation. The transition is step-function-like: the jump from 0 paths (eligibility traces, fails) to any $k{>}0$ spans two orders of magnitude; from $k{=}4$ to $k{=}64$ is noise. The redundancy extends to transformers via sparse gradient transport (outperforming dense at 50\% head sparsity; borderline at 33\%) and to real primate neural data ($80\%$ recovery on cross-session electrode drift), with architecture-dependent thresholds explained by unit specialization.

Our contribution is empirical, not algorithmic. We do not propose sparse Jacobian propagation as a new method to be optimized. The finding is that gradient transport in neural networks has a structural property (massive redundancy) that makes careful optimization unnecessary. In RNNs, the mechanism is near-isotropy of the Jacobian's singular value spectrum. The same redundancy appears in LSTMs and transformers, though with different mechanisms and higher thresholds (${\sim}6\%$ of neurons for RNNs, ${\sim}33\%$ of attention heads for transformers). For RNNs, this reframes the RTRL approximation problem: the bottleneck is having \emph{any} Jacobian propagation at all, not having the \emph{right} propagation. For transformers, where heads are specialized, selection matters more ($+$24.3pp Fisher oracle--anti gap), but a redundancy threshold still exists. One scope condition: the redundancy holds under continuous, dense error signals. On tasks with sparse supervision (copy tasks, adding problems), Jacobian propagation accumulates numerical drift and all RTRL variants degrade. Within the dense error signal regime where RTRL has practical value, most of the Jacobian is unnecessary.

\section{Background}
\label{sec:background}

\subsection{Real-Time Recurrent Learning}
\label{sec:rtrl}

Consider a vanilla RNN with hidden state $h_t \in \mathbb{R}^n$, input $x_t$, and output $y_t$:
\begin{align}
    h_t &= \tanh(W_{hh}\, h_{t-1} + W_{ih}\, x_t + b_h) \label{eq:rnn} \\
    y_t &= W_{\text{out}}\, h_t + b_{\text{out}} \label{eq:output}
\end{align}
RTRL \citep{williams1989rtrl} computes exact gradients in forward mode by maintaining the Jacobian tensor $J_t[i, j, k] = \partial h_t^{(i)} / \partial \theta_{jk}$, where $\theta$ denotes the recurrent parameters ($W_{hh}$, $W_{ih}$, $b_h$). The Jacobian evolves as:
\begin{equation}
    J_t = D_t \left( W_{hh}\, J_{t-1} + B_t \right)
    \label{eq:rtrl}
\end{equation}
where $D_t = \text{diag}(1 - h_t^2)$ is the derivative of $\tanh$ and $B_t$ contains the immediate (direct) derivatives of $h_t$ with respect to $\theta$: for $W_{hh}$, row $i$ of $B_t$ is $h_{t-1}^\top$ in the block corresponding to weight row $i$, and zero elsewhere.

The term $W_{hh}\, J_{t-1}$ is the \emph{Jacobian propagation term}. It captures how sensitivity to parameters ripples through recurrent connections over time, the mechanism by which RTRL assigns credit across time steps. The cost of computing this term dominates: the Jacobian tensor has $n \times n \times |\theta_{\text{rec}}|$ entries, and the matrix-tensor contraction with $W_{hh}$ costs $O(n^4)$ per step for the recurrent weight group ($n^2$ parameters, each requiring $O(n^2)$ work).

Given $J_t$, the gradient of a per-step loss $L_t = \ell(y_t, \text{target}_t)$ is:
\begin{equation}
    \frac{\partial L_t}{\partial \theta} = \frac{\partial L_t}{\partial y_t} \, W_{\text{out}} \, J_t
    \label{eq:rtrl_grad}
\end{equation}
This gradient is exact, online (computable at each step without stored history), and local in time (depends only on current quantities). The price is the $O(n^4)$ Jacobian update.

\subsection{Eligibility Traces}
\label{sec:traces}

Eligibility traces \citep{murray2019rflo, bellec2020eprop} approximate RTRL by dropping the Jacobian propagation term entirely:
\begin{equation}
    \tilde{J}_t = D_t \, B_t
    \label{eq:traces}
\end{equation}
or, with a decay factor $\lambda$:
\begin{equation}
    \tilde{J}_t = \lambda \, D_t \, \tilde{J}_{t-1} + D_t \, B_t
    \label{eq:traces_decay}
\end{equation}
In both cases, the $W_{hh}\, J_{t-1}$ term is absent. This reduces cost to $O(n^2)$ per step but assumes that each weight only affects the hidden state at the current time step. The approximation discards all sensitivity that propagates through recurrent connections, precisely the mechanism needed for temporal credit assignment.

This is not a minor omission. Eligibility traces are equivalent to RTRL in a feedforward network (where $W_{hh} = 0$). In a recurrent network, they retain only the ``direct'' component of the gradient and lose the ``recurrent'' component that carries information about how parameter changes at time $t$ affect hidden states at times $t+1, t+2, \ldots$ It is this recurrent component that enables adaptation to distributional shifts, as we demonstrate empirically in Section~\ref{sec:experiments}.

\subsection{Prior RTRL Approximations}
\label{sec:prior}

Several methods approximate the Jacobian propagation term rather than dropping it entirely. Table~\ref{tab:prior} summarizes prior work.

\begin{table}[h]
\centering
\caption{RTRL approximation methods. All propose structured approximations with specific design tradeoffs. None tests whether unstructured random sparsification suffices.}
\label{tab:prior}
\begin{tabular}{llll}
\toprule
Method & Mechanism & Cost & Constraint \\
\midrule
UORO \citep{tallec2017uoro} & Rank-1 compression & $O(n^2)$ & High variance \\
SnAp \citep{menick2021snap} & Graph-structured sparsity & $O(n^2 s)$ & Weight-sparse RNN \\
OK \citep{benzing2019ok} & Kronecker-sum approx & $O(n^3)$ & None \\
Element-wise \citep{irie2024elstm} & Kill inter-neuron recurrence & $O(n^2)$ & Architectural change \\
RFLO \citep{murray2019rflo} & Drop Jacobian term & $O(n^2)$ & $=$ eligibility traces \\
\bottomrule
\end{tabular}
\end{table}

The closest prior work is SnAp \citep{menick2021snap}, which also sparsifies the Jacobian propagation but uses graph-structured sparsity in weight-sparse RNNs. We compare in detail in Section~\ref{sec:related}.

Every method in Table~\ref{tab:prior} makes a specific structural choice: which rank (UORO), which graph distance (SnAp), which Kronecker structure (OK), which neurons to decouple (Irie). We test whether any of these choices matter. They largely do not.

\section{Method}
\label{sec:method}

\subsection{Sparse Jacobian Propagation}
\label{sec:sparse}

The full RTRL update (Eq.~\ref{eq:rtrl}) propagates sensitivity through all $n$ recurrent connections for each neuron $i$:
\begin{equation}
    J_t[i, :, :] = (1 - h_t^{(i)2}) \left( \sum_{l=1}^{n} W_{hh}[i,l] \cdot J_{t-1}[l, :, :] + B_t[i, :, :] \right)
    \label{eq:full_prop}
\end{equation}

We restrict the sum to a fixed random subset $\mathcal{S}_i \subset \{1, \ldots, n\}$ of size $k$ for each neuron $i$:
\begin{equation}
    \hat{J}_t[i, :, :] = (1 - h_t^{(i)2}) \left( \sum_{l \in \mathcal{S}_i} W_{hh}[i,l] \cdot \hat{J}_{t-1}[l, :, :] + B_t[i, :, :] \right)
    \label{eq:sparse_prop}
\end{equation}

The immediate derivative term $B_t$ is unchanged: every neuron retains full information about how its own parameters directly affect its own activation. Only the \emph{propagation} of sensitivity through recurrent connections is sparsified. The network weights $W_{hh}$ remain fully dense; the network architecture and forward dynamics are identical to a standard RNN. Only the gradient computation is approximate.

\paragraph{Implementation.} In practice, we construct a binary mask $M \in \{0, 1\}^{n \times n}$ where $M[i, l] = 1$ iff $l \in \mathcal{S}_i$, and compute the masked Jacobian update as:
\begin{equation}
    \hat{J}_t = D_t \left( (W_{hh} \odot M)\, \hat{J}_{t-1} + B_t \right)
    \label{eq:masked}
\end{equation}
where $\odot$ denotes elementwise multiplication. This replaces the $W_{hh} J_{t-1}$ contraction in the standard einsum with $(W_{hh} \odot M) J_{t-1}$, zeroing out contributions from neurons outside each neuron's selected subset.

\paragraph{Subset construction.} Our default selection arranges hidden units on an arbitrary ring and assigns each neuron $i$ the $k/2$ neighbors on each side. Since $W_{hh}$ is dense, this ring ordering has no relation to the network's learned connectivity; it is a random assignment. We verify this by testing four additional selection strategies in Section~\ref{sec:exp3}, including random selection and adversarial (worst-case) selection.

\subsection{Cost Analysis}
\label{sec:cost}

The computational bottleneck in RTRL is the Jacobian propagation: the contraction of $W_{hh}$ (an $n \times n$ matrix) with $J_{t-1}$ (an $n \times n \times p$ tensor, where $p$ is the number of parameters in the corresponding group). For $W_{hh}$ itself, $p = n^2$, so the contraction costs $O(n \cdot n \cdot n^2) = O(n^4)$ per step.

With sparse propagation, each row $i$ of the contraction sums over $k$ terms instead of $n$. The cost reduces to $O(k \cdot n^3)$ per step, a factor of $n/k$ cheaper than full RTRL. For $k{=}4$ and $n{=}64$, this is a 16$\times$ reduction; for $k{=}4$ and $n{=}256$, a 64$\times$ reduction. The savings scale linearly with network size because $k$ remains fixed while $n$ grows.

\begin{table}[h]
\centering
\caption{Per-step computational cost for RTRL variants.}
\label{tab:cost}
\begin{tabular}{lll}
\toprule
Method & Cost per step & Notes \\
\midrule
Full RTRL & $O(n^4)$ & Exact forward-mode gradients \\
Sparse RTRL ($k$ neighbors) & $O(k \cdot n^3)$ & $k \ll n$; dense weights \\
Eligibility traces & $O(n^2)$ & No Jacobian propagation \\
UORO & $O(n^2)$ & Rank-1, high variance \\
SnAp & $O(n^2 \cdot s)$ & $s$ = propagation steps; weight-sparse \\
\bottomrule
\end{tabular}
\end{table}

\subsection{Evaluation Protocol: Online Adaptation}
\label{sec:protocol}

We evaluate on a task that directly tests temporal credit assignment: \emph{online next-step prediction with mid-stream distribution shifts}. The network trains continuously on a streaming signal, receiving one input per time step and updating parameters immediately. At a predetermined point, the data distribution changes (e.g., a frequency shift in a sine wave, a parameter change in a dynamical system). We measure prediction quality before and after the shift.

This evaluation differs from standard RTRL benchmarks in the literature. Copy tasks \citep{hochreiter1997lstm}, character-level language modeling \citep{benzing2019ok}, and adding problems \citep{hochreiter1997lstm} test whether a method can learn a fixed task. Our protocol tests whether a method can \emph{adapt} when the task changes, a more direct test of the temporal credit assignment mechanism, since adaptation requires the gradient to carry information about how the current dynamics differ from what the network was trained on.

\paragraph{Gap recovery metric.} To compare methods across tasks with different MSE scales, we define \emph{gap recovery} on a logarithmic scale. Let $\text{MSE}_0$ denote the post-shift MSE with eligibility traces ($k{=}0$, the floor: no Jacobian propagation, no adaptation) and $\text{MSE}_n$ denote the post-shift MSE with full RTRL ($k{=}n$, the ceiling: exact Jacobian, full adaptation). For a sparse variant with $k$ neighbors:
\begin{equation}
    \text{Recovery}(k) = \frac{\log \text{MSE}_0 - \log \text{MSE}_k}{\log \text{MSE}_0 - \log \text{MSE}_n} \times 100\%
    \label{eq:recovery}
\end{equation}
A recovery of 0\% means the sparse variant is no better than eligibility traces; 100\% means it matches full RTRL. Intuitively, 50\% recovery means the sparse variant achieves the geometric mean of the trace and full-RTRL error levels. Values above 100\% indicate the sparse variant outperforms full RTRL, possible when sparsification acts as implicit regularization (Section~\ref{sec:exp2}).

\section{Experiments}
\label{sec:experiments}

We test the redundancy of gradient transport through nine experiments, each answering one question in a logical chain: (1) does any partial Jacobian propagation recover adaptation? (2) does this hold on complex, chaotic dynamics and larger networks? (3) does the \emph{choice} of which paths to keep matter? (4) where does it break? (5) does the redundancy extend to gated architectures? (6) does the absolute-$k$ scaling law hold at larger network sizes? (7) does an analogous redundancy extend to transformers? (8) does head selection matter in transformers? (9) does the finding transfer to real-world neural data?

Experiments 1--6 use the same core setup: an RNN with explicit parameters trained online with Adam ($\text{lr}=0.001$, $\beta_1=0.9$, $\beta_2=0.999$). Experiments 1--4 use a vanilla tanh RNN with dense $W_{hh}$; Experiment 5 uses an LSTM; Experiment 7 uses a vision transformer on continual image classification. We evaluate on \emph{online adaptation to distribution shifts}: the data stream changes mid-sequence (or mid-task-sequence for transformers), and we measure how well each method tracks the new distribution. This evaluation angle is distinct from standard RTRL benchmarks (which test convergence on a fixed task) and directly tests the credit-assignment mechanism we are studying. RNN experiments run on CPU in minutes; transformer experiments run on GPU. Full implementation details (hyperparameters, architectures, training schedules) are provided alongside each experiment.

\paragraph{Metric.} We report gap recovery (Eq.~\ref{eq:recovery}) throughout; values above 100\% indicate the sparse variant outperforms full RTRL.

\paragraph{Implementation validation.} Our RTRL and eligibility trace implementations were validated against the independent reference implementation of \citet{marschall2020unified}, producing exact numerical agreement (zero difference) on both full RTRL and RFLO baselines across 2000 online steps with shared weight initialization.

\subsection{Experiment 1: The Jacobian is massively redundant}
\label{sec:exp1}

How much of the recurrent Jacobian do you actually need? We test this on the simplest possible task: a 1D sine wave whose frequency shifts from 0.1 to 0.3 at $t{=}1000$ in a 2000-step online stream. A vanilla RNN ($n{=}64$) performs next-step prediction. We sweep the number of Jacobian propagation paths: $k \in \{0, 4, 8, 16, 32, 64\}$, where $k{=}0$ is eligibility traces and $k{=}64$ is full RTRL. Paths are selected as $k/2$ contiguous neighbors on each side of an arbitrary ring ordering over hidden units.

The result is a step function. Table~\ref{tab:exp1} shows that $k{=}4$ (6\% of hidden units) recovers $84 \pm 6\%$ of full RTRL's adaptation ability across 5 seeds. The jump from $k{=}0$ to $k{=}4$ spans two orders of magnitude in post-shift MSE (0.23 $\to$ 0.0015). The jump from $k{=}4$ to $k{=}64$ is noise.

\begin{table}[h]
\centering
\caption{Post-shift MSE and gap recovery on sine frequency shift ($n{=}64$, 5 seeds, mean $\pm$ s.d.). The jump from $k{=}0$ to any $k \geq 4$ spans two orders of magnitude; further increasing $k$ adds no systematic improvement.}
\label{tab:exp1}
\begin{tabular}{lccc}
\toprule
$k$ & Fraction & Post-shift MSE & Recovery (\%) \\
\midrule
0 (traces) & 0\% & $0.23 \pm 0.15$ & 0 \\
4 & 6\% & $0.0015 \pm 0.0005$ & $84 \pm 6$ \\
8 & 12.5\% & $0.0016 \pm 0.0005$ & $82 \pm 5$ \\
16 & 25\% & $0.0017 \pm 0.0005$ & $82 \pm 6$ \\
32 & 50\% & $0.0012 \pm 0.0002$ & $87 \pm 3$ \\
64 (full) & 100\% & $0.0006 \pm 0.0000$ & 100 \\
\midrule
BPTT $w{=}1$ (ref) & --- & $0.0015 \pm 0.0007$ & --- \\
\bottomrule
\end{tabular}
\end{table}

\begin{figure}[t]
\centering
\includegraphics[width=\textwidth]{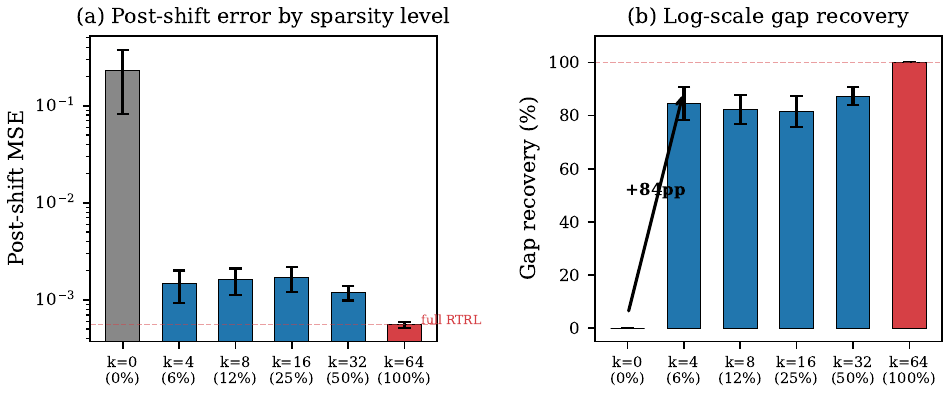}
\caption{The Jacobian is massively redundant. (a) Post-shift MSE drops by two orders of magnitude from $k{=}0$ (eligibility traces) to $k{=}4$ (6\% of paths), then stays flat through $k{=}64$ (full RTRL). (b) Log-scale gap recovery is a step function: any $k \geq 4$ recovers $82$--$87\%$ of full RTRL. Error bars: $\pm 1$ s.d.\ across 5 seeds.}
\label{fig:recovery_step}
\end{figure}

Two things stand out. First, the recovery is flat across $k{=}4$ to $k{=}32$: all achieve $82$--$87\%$ log-scale recovery with overlapping confidence intervals ($\pm 3$--$6\%$). In absolute terms, all partial variants achieve post-shift MSE of $0.001$--$0.002$, within a factor of 3 of full RTRL, while $k{=}0$ is two orders of magnitude worse. The Jacobian information needed for online adaptation is present in any small random subset of the propagation paths. Second, the BPTT reference ($0.0015 \pm 0.0007$) uses per-step window-1 backpropagation and achieves MSE comparable to $k{=}4$; both are limited by per-step credit assignment, but even this crude Jacobian propagation captures the relevant signal.

\paragraph{Comparison with truncated BPTT.} In a separate ablation (3 seeds, learning rates selected per-window via sweep), we compare sparse RTRL against truncated BPTT with windows $w \in \{1, 10, 50\}$. Sparse RTRL ($k{=}4$) achieves post-shift MSE of $0.0011 \pm 0.0001$, comparable to or better than BPTT $w{=}1$ ($0.0014 \pm 0.0009$) and BPTT $w{=}10$ ($0.0016 \pm 0.0019$), while operating fully online with $O(1)$ temporal memory. BPTT $w{=}50$ fails to converge properly (pre-shift MSE $0.021$) despite learning rates being swept for all window sizes including $w{=}50$. This illustrates truncated BPTT's sensitivity to the window--learning-rate interaction, a tuning problem sparse RTRL avoids entirely. Sparse RTRL also exhibits substantially lower variance across seeds than either BPTT variant.

\subsection{Experiment 2: Scaling to chaotic dynamics}
\label{sec:exp2}

Does this redundancy survive when the dynamics are genuinely complex? A sine wave is a single-frequency oscillator. We now test on the Lorenz attractor, a 3D chaotic system with nonlinear coupling between state variables. The RNN ($n{=}64$, $n{=}128$) predicts the next state of the Lorenz system integrated via RK4 ($dt{=}0.01$). At $t{=}2000$, the Lorenz parameter $\rho$ shifts from 28 to 20, qualitatively changing the attractor geometry. We also test multi-regime sine waves (4 frequency regimes, 3 shift points) to evaluate repeated adaptation.

The redundancy holds. On chaotic dynamics, sparse Jacobian propagation is actually \emph{more numerically stable} than full RTRL. We did not expect an approximation to outperform the exact method. Table~\ref{tab:lorenz} shows post-shift MSE for $k{=}4$ and full RTRL across 5 seeds at both $n{=}64$ and $n{=}128$. The main result: $k{=}4$ produces tight, stable predictions across all seeds (coefficient of variation 8--13\%), while full RTRL diverges on a seed-dependent fraction (CV 88--161\%). At $n{=}64$, full RTRL fails on 2 of 5 seeds (post-shift MSE exceeding traces); at $n{=}128$, it fails on 3 of 5. Sparse propagation implicitly regularizes the exponentially growing chaotic Jacobian (Section~\ref{sec:discussion}).

\begin{table}[h]
\centering
\caption{Post-shift MSE on Lorenz attractor (5 seeds). $k{=}4$ is stable across all seeds; full RTRL diverges on chaotic dynamics. Bold values indicate failure (MSE $>$3$\times$ the $k{=}4$ mean).}
\label{tab:lorenz}
\begin{tabular}{lcccc}
\toprule
 & \multicolumn{2}{c}{$n{=}64$} & \multicolumn{2}{c}{$n{=}128$} \\
\cmidrule(lr){2-3} \cmidrule(lr){4-5}
Seed & $k{=}4$ & Full RTRL & $k{=}4$ & Full RTRL \\
\midrule
42 & 0.015 & 0.015 & 0.016 & \textbf{0.067} \\
123 & 0.016 & 0.011 & 0.015 & 0.021 \\
7 & 0.018 & \textbf{0.046} & 0.015 & \textbf{0.876} \\
2024 & 0.012 & 0.013 & 0.018 & \textbf{0.063} \\
31337 & 0.014 & \textbf{0.093} & 0.018 & 0.012 \\
\midrule
Mean $\pm$ s.d. & $0.015 \pm 0.002$ & $0.036 \pm 0.035$ & $0.016 \pm 0.001$ & $0.208 \pm 0.374$ \\
\bottomrule
\end{tabular}
\end{table}

\begin{figure}[t]
\centering
\includegraphics[width=\textwidth]{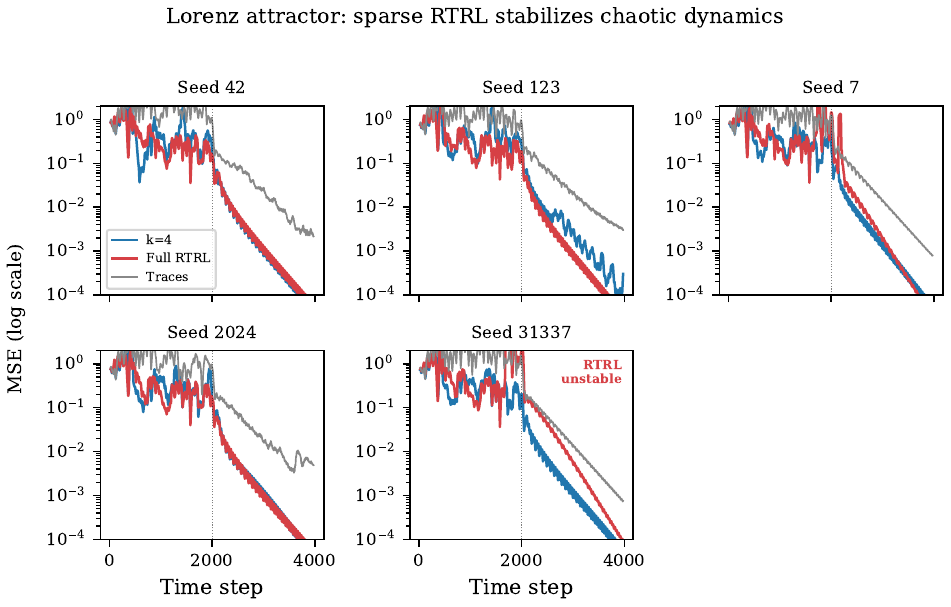}
\caption{Sparse RTRL is more stable than full RTRL on chaotic dynamics. Per-seed loss curves on the Lorenz attractor ($n{=}64$, 5 seeds). Full RTRL (red) degrades on 2 of 5 seeds; $k{=}4$ (blue) is stable on all seeds. Gray: eligibility traces ($k{=}0$). Vertical dotted line: parameter shift at $t{=}2000$.}
\label{fig:lorenz_stability}
\end{figure}

The scaling result matters: $k{=}4$ at $n{=}128$ (3.1\% of hidden units) achieves post-shift MSE of $0.016 \pm 0.001$, nearly identical to $k{=}4$ at $n{=}64$ ($0.015 \pm 0.002$). The required number of propagation paths does not grow with network size; it is the absolute count $k$, not the fraction $k/n$, that determines performance. We confirm this scaling law up to $n{=}256$ in Experiment~6 (Section~\ref{sec:exp6}).

On multi-regime sine (Table~\ref{tab:multisine}), $k{=}4$ recovers $80 \pm 12\%$ across 5 seeds at the first shift point, with no anomalous seeds or RTRL instability (multi-sine is non-chaotic).

\begin{table}[h]
\centering
\caption{Post-shift MSE and gap recovery on multi-regime sine at first shift point ($n{=}64$, 5 seeds). Unlike Lorenz, full RTRL is stable on non-chaotic dynamics.}
\label{tab:multisine}
\begin{tabular}{lcccc}
\toprule
Seed & Traces & $k{=}4$ & Full RTRL & Recovery (\%) \\
\midrule
42 & 0.100 & 0.0072 & 0.0022 & 69 \\
123 & 0.178 & 0.0037 & 0.0019 & 85 \\
7 & 0.202 & 0.0028 & 0.0020 & 93 \\
2024 & 0.040 & 0.0057 & 0.0020 & 65 \\
31337 & 0.100 & 0.0035 & 0.0023 & 89 \\
\midrule
Mean $\pm$ s.d. & & $0.0046 \pm 0.0016$ & $0.0021 \pm 0.0001$ & $80 \pm 12$ \\
\bottomrule
\end{tabular}
\end{table}

\paragraph{Comparison with UORO.} We compare against UORO \citep{tallec2017uoro}, a published rank-1 Jacobian approximation. The comparison reveals a task-dependent tradeoff. On Lorenz, UORO achieves the best post-shift MSE of any method ($0.0025 \pm 0.0006$ across 5 seeds), outperforming both $k{=}4$ ($0.015 \pm 0.002$) and full RTRL ($0.036 \pm 0.035$). UORO's rank-1 compression acts as strong regularization on the exponentially growing chaotic Jacobian. On multi-regime sine (non-chaotic), UORO's compression destroys useful gradient information: post-shift MSE of $0.049 \pm 0.005$ (5 seeds), an order of magnitude worse than $k{=}4$ ($0.005 \pm 0.002$). Sparse Jacobian propagation is the more robust general-purpose approach; UORO has a niche advantage specific to chaotic dynamics where the exact Jacobian is pathological.

\subsection{Experiment 3: Selection does not matter}
\label{sec:exp3}

We have shown that a random 6\% of Jacobian paths suffices. But does the \emph{choice} of which paths to keep matter? If an oracle that selects the ``best'' paths significantly outperforms random selection, there is headroom for learned selection mechanisms. If even deliberately adversarial selection works, the redundancy is so complete that optimization of the selection is unnecessary.

We test five selection strategies, all at $k{=}4$ on $n{=}64$:
\begin{itemize}
    \item Ring: contiguous neighbors on an arbitrary ring (baseline from Experiments 1--2)
    \item Random: random $k$ per neuron, fixed at initialization
    \item Oracle ($\text{top}~|W_{hh}|$): strongest recurrent connections by weight magnitude, recomputed every 200 steps
    \item Anti-oracle ($\text{bottom}~|W_{hh}|$): \emph{weakest} recurrent connections, deliberately the worst choice
    \item Dynamic oracle: top $k$ by $|W_{hh}[i,l]| \cdot \|J[l,:,:]\|$, recomputed per step (uses gradient information)
\end{itemize}

The result is unambiguous: selection does not matter. Table~\ref{tab:selection} shows post-shift MSE for all five strategies across 5 seeds on both tasks. On sine, all strategies achieve post-shift MSE of $0.0014$--$0.0019$ (mean across seeds), with no strategy systematically dominating. On Lorenz, the dynamic Jacobian oracle is slightly \emph{worst} ($0.020 \pm 0.006$). Per-step reselection based on the chaotic Jacobian adds noise rather than signal.

\begin{table}[h]
\centering
\caption{Post-shift MSE for five selection strategies, all at $k{=}4$ ($n{=}64$, 5 seeds, mean $\pm$ s.d.). No strategy is systematically better; the ranking changes across seeds.}
\label{tab:selection}
\begin{tabular}{lccccc}
\toprule
Task & Ring & Random & Oracle & Anti-oracle & Dynamic \\
\midrule
Sine & $\mathbf{.0015 \pm .0005}$ & $.0016 \pm .0006$ & $.0016 \pm .0004$ & $\mathbf{.0014 \pm .0008}$ & $.0019 \pm .0004$ \\
Lorenz & $\mathbf{.0149 \pm .0019}$ & $.0156 \pm .0006$ & $.0155 \pm .0021$ & $.0164 \pm .0010$ & $.0199 \pm .0063$ \\
\bottomrule
\end{tabular}
\end{table}

Across 50 strategy-seed combinations (5 strategies $\times$ 5 seeds $\times$ 2 tasks), no strategy achieves a consistent advantage. The information in the Jacobian is so redundant that any non-degenerate subset above a minimal threshold (random, structured, or adversarial) captures enough for adaptation. The design choice that matters is not \emph{which} paths to keep, but \emph{how many}, and the answer from Experiment 1 is: surprisingly few.

\subsection{Experiment 4: When forward-mode breaks down}
\label{sec:exp4}

All experiments so far use tasks with continuous error signals: the network receives prediction error at every time step, and the Jacobian is grounded by gradient updates throughout the sequence. What happens when error is sparse?

We test two standard RTRL benchmarks: the \emph{copy task} \citep{hochreiter1997lstm} (memorize and reproduce a symbol sequence across a delay, cross-entropy loss only on output phase) and the \emph{adding problem} \citep{hochreiter1997lstm} (identify two marked values in a 50-step sequence, MSE loss only at the final step). Both tasks provide error signal at a fraction of time steps, leaving the Jacobian to propagate for extended periods without corrective feedback.

Full RTRL ($k{=}64$) is the \emph{worst} method on both tasks, worse than BPTT, worse than eligibility traces. Table~\ref{tab:scope} shows the numbers. On the adding problem, full RTRL achieves MSE of 0.289, compared to the trivial baseline of 0.167 (predicting the expected sum of two uniform random variables).

\begin{table}[h]
\centering
\caption{Standard RTRL benchmarks with vanilla tanh RNN ($n{=}64$). No method succeeds; full RTRL is worst on both tasks. The architecture cannot solve these tasks regardless of training method.}
\label{tab:scope}
\begin{tabular}{lcc}
\toprule
Method & Copy T=5 (accuracy $\uparrow$) & Adding T=50 (MSE $\downarrow$) \\
\midrule
Trivial baseline & 0.125 (chance) & 0.167 (predict mean) \\
\midrule
BPTT & 0.289 & 0.189 \\
$k{=}0$ (traces) & 0.239 & 0.284 \\
$k{=}4$ & 0.239 & 0.281 \\
$k{=}8$ & 0.292 & 0.278 \\
$k{=}64$ (full RTRL) & 0.130 & \textbf{0.289} (worst) \\
\bottomrule
\end{tabular}
\end{table}

Two distinct phenomena contribute to this result. First, the vanilla tanh RNN lacks the gating mechanisms needed for discrete memory tasks; these benchmarks were designed to motivate LSTMs \citep{hochreiter1997lstm}, and published methods that succeed on them use GRU or LSTM architectures \citep{menick2021snap, tallec2017uoro}. No gradient method can overcome an architecture that cannot maintain discrete states through tanh contraction. This makes the experiment uninformative for comparing gradient approximations: there is no gap between traces and full RTRL to measure.

Second, and more interesting: full RTRL is the \emph{worst} method, underperforming even eligibility traces. Jacobian propagation without continuous error feedback is actively harmful. Without per-step gradient corrections, the Jacobian tensor accumulates numerical drift over $T$ steps of ungrounded propagation, producing gradient estimates that are worse than the crude trace approximation. This predicts that RTRL-based methods will struggle on any task with sparse or delayed reward signals, a scope condition for the entire family of forward-mode approaches, not only for our sparse variant.

\subsection{Experiment 5: The redundancy extends to LSTMs}
\label{sec:exp5}

Experiments 1--3 establish Jacobian redundancy in vanilla tanh RNNs, where the recurrent weight matrix $W_{hh}$ is the sole path for temporal credit assignment. LSTMs provide a qualitatively different architecture: the cell state acts as a nearly linear highway through time, with gating mechanisms controlling information flow. Does the cell state highway eliminate the need for Jacobian propagation through the gates, or does the redundancy persist?

We test an LSTM ($n{=}64$) with explicit gate Jacobians for RTRL. The cell state Jacobian is element-wise (diagonal) and always computed exactly; it requires no sparsification. We sparsify only the inter-neuron gate Jacobians: how gate activations at one neuron propagate sensitivity through connections to other neurons. This is the LSTM analogue of sparsifying $W_{hh}$ in the vanilla RNN. The task is sine frequency shift (0.1 $\to$ 0.3), with a longer sequence ($T{=}4000$, shift at $t{=}2000$) to ensure the LSTM converges before the shift.\footnote{An initial run with $T{=}2000$ (shift at $t{=}1000$) produced an inverted gap where traces appeared to match full RTRL. This was a convergence artifact: with insufficient pre-training, all methods were still learning the initial distribution when the shift occurred, and the post-shift comparison was meaningless. Doubling the pre-training time resolved the artifact and revealed the pattern reported here. We include this as a cautionary note: verifying pre-shift convergence is essential before interpreting adaptation results.}

\begin{table}[h]
\centering
\caption{LSTM gate Jacobian sparsification on sine frequency shift ($n{=}64$, $T{=}4000$, shift at $t{=}2000$, 5 seeds). The vanilla RNN pattern holds: traces fail, $k{=}4$ recovers near-full RTRL performance. Seed 123 excluded from recovery (inverted gap; traces converge better than full RTRL on that seed).}
\label{tab:lstm}
\begin{tabular}{lcccc}
\toprule
Seed & Traces & $k{=}4$ & Full RTRL & Notes \\
\midrule
42 & 0.508 & 0.004 & 0.002 & Clean \\
123 & 0.002 & 0.001 & 0.015 & Inverted \\
7 & 0.548 & 0.003 & 0.015 & $k{=}4$ beats full \\
2024 & 0.220 & 0.001 & 0.0002 & Clean \\
31337 & 0.090 & 0.001 & 0.004 & $k{=}4$ beats full \\
\midrule
Mean $\pm$ s.d. & & $0.002 \pm 0.001$ & $0.007 \pm 0.006$ & \\
\bottomrule
\end{tabular}
\end{table}

The result matches the vanilla RNN pattern (Table~\ref{tab:lstm}). Eligibility traces fail on 4 of 5 seeds (post-shift MSE $0.09$--$0.55$), two orders of magnitude worse than $k{=}4$.\footnote{Seed 123 is anomalous: traces achieve 0.002 post-shift (better than full RTRL at 0.015), producing an inverted gap. Neither method converged well pre-shift on this seed; the trace approximation happened to suit the post-shift distribution.} Sparse gate Jacobian propagation with $k{=}4$ achieves post-shift MSE of $0.002 \pm 0.001$ across all 5 seeds, comparable to or better than full RTRL ($0.007 \pm 0.006$). On 2 of 5 seeds, $k{=}4$ outperforms full RTRL, the same sparse regularization effect observed on the Lorenz attractor.

The cell state highway does not eliminate the need for gate Jacobian propagation. Despite providing a linear gradient path through time, the LSTM still requires inter-neuron sensitivity information flowing through the gates, and this information exhibits the same redundancy as in vanilla RNNs. As with Lorenz, $k{=}4$ is more numerically stable than full RTRL: CV of 69\% vs 88\%.

We also tested the LSTM on the Lorenz attractor (with extended pre-training, $T{=}8000$, shift at $t{=}4000$). LSTM RTRL did not converge to acceptable pre-shift performance on chaotic dynamics: pre-shift MSE remained 0.08--0.90 across all methods and seeds, compared to 0.01--0.05 for the vanilla RNN. This affects full RTRL as well as sparse variants, and is consistent with the theoretical result of \citet{mikhaeil2022chaos} that loss gradients of recurrent networks producing chaotic dynamics always diverge, regardless of architecture (their result covers all first-order Markovian recursive maps, including LSTMs and GRUs). We conjecture that the LSTM's four-gate structure compounds rather than ameliorates this instability. The LSTM redundancy finding is therefore established on smooth dynamics (sine) but inconclusive on chaotic dynamics.

This extends the redundancy finding from architecture-specific to architecture-general on smooth dynamics. The bottleneck for online adaptation is having \emph{any} Jacobian propagation through the gates, not having the right propagation, and not the cell state highway alone.

\subsection{Experiment 6: Scaling to $n{=}256$}
\label{sec:exp6}

Experiments 1--2 established the absolute-$k$ scaling law at $n{=}64$ and $n{=}128$: the number of Jacobian paths needed for adaptation does not grow with network size. We now test whether this holds at $n{=}256$, where $k{=}4$ represents just 1.6\% of neurons, and at $n{=}512$ (0.8\%), where the computational savings of sparse RTRL become substantial (64$\times$ and 128$\times$ respectively).

\begin{table}[h]
\centering
\caption{Scale test results at $n{=}256$ (sine frequency shift, 5 seeds). $k{=}4$ at 1.6\% recovers $78 \pm 20\%$ of the log-space gap. Seed 31337: $k{=}4$ outperforms full RTRL (recovery $>$100\%), another stability win consistent with the Lorenz findings.}
\label{tab:scale}
\begin{tabular}{lccc}
\toprule
Seed & $k{=}4$ post-shift MSE & $k{=}256$ (full) & Recovery (\%) \\
\midrule
42 & 0.0036 & 0.0003 & 66 \\
123 & 0.0054 & 0.0003 & 62 \\
7 & 0.0036 & 0.0010 & 80 \\
2024 & 0.0034 & 0.0003 & 68 \\
31337 & 0.0032 & 0.0065 & 116 \\
\midrule
Mean $\pm$ s.d. & $0.0038 \pm 0.0008$ & & $78 \pm 20$ \\
\bottomrule
\end{tabular}
\end{table}

At $n{=}256$, sparse propagation ($k{=}4$) achieves consistent post-shift MSE of $0.0038 \pm 0.0008$ across all 5 seeds (Table~\ref{tab:scale}). Full RTRL shows 22$\times$ variance across seeds ($0.0003$--$0.0065$), with seed 31337 partially failing to adapt, consistent with the instability pattern observed on chaotic dynamics at smaller scales. The recovery variance ($78 \pm 20\%$) inherits this instability from the full RTRL ceiling; $k{=}4$ absolute performance is scale-stable. Traces remain at ${\sim}0.51$ across all seeds (two orders of magnitude worse than $k{=}4$). Spectral analysis at $n{=}256$ (Table~\ref{tab:spectral}) confirms that the Jacobian remains near-isotropic ($r_{95}/n = 92.5\%$, condition number $2.97 \pm 0.33$, actually better conditioned than at $n{=}64$ or $n{=}128$). The recovery variance increase despite improving isotropy suggests that numerical precision, not spectral structure, is the limiting factor at larger scales.

The scaling law across all tested network sizes is:

\begin{center}
\begin{tabular}{lcccccc}
\toprule
$n$ & $k{=}4$ fraction & $k{=}4$ MSE & Recovery & Seeds & Task \\
\midrule
64 & 6.25\% & $0.0015 \pm 0.0005$ & $84 \pm 6\%$ & 5 & Sine \\
128 & 3.13\% & $0.016 \pm 0.001$ & stable$^\dagger$ & 5 & Lorenz \\
256 & 1.56\% & $0.0038 \pm 0.0008$ & $78 \pm 20\%$ & 5 & Sine \\
\bottomrule
\multicolumn{6}{l}{\footnotesize $^\dagger$Full RTRL diverges on 2/5 seeds; $k{=}4$ is stable on all 5 (CV 13\%).}
\end{tabular}
\end{center}

The fraction $k/n$ decreases by 4$\times$ from $n{=}64$ to $n{=}256$, while $k{=}4$ absolute performance degrades modestly ($0.0015 \to 0.0038$). Full RTRL degrades catastrophically: from stable at $n{=}64$ to 22$\times$ seed-dependent variance at $n{=}256$. The practical reliability gap between sparse and full propagation \emph{widens} as $n$ grows.

At $n{=}512$, two distinct failure modes emerge. First, \emph{numerical divergence}: on sine, full RTRL ($k{=}512$) diverges on 1 of 2 tested seeds (post-shift MSE 1.8, far worse than eligibility traces and BPTT), while converging normally on the other. The Jacobian tensor at this scale ($(512, 512, 512)$, comprising 134 million floating-point entries) accumulates numerical errors through 2000 steps of repeated matrix-tensor contractions, causing the sensitivity estimates to blow up. Second, \emph{overparameterization}: a 512-dimensional network has far more capacity than these tasks require, so local per-step updates suffice without temporal credit assignment. On Lorenz at $n{=}512$ (with full RTRL skipped due to computational cost), eligibility traces achieve near-perfect performance on some seeds (post-shift MSE $3 \times 10^{-6}$ on seed 123), while Jacobian propagation actively hurts, injecting noise from a numerically unstable tensor into a problem that does not require it. These two failure modes bound the absolute-$k$ scaling law from above: it holds up to the point where either the Jacobian becomes too large to propagate stably, or the task no longer requires temporal credit assignment due to overparameterization. This establishes $n \approx 256$--$512$ as the practical ceiling for forward-mode Jacobian propagation with standard floating-point arithmetic on these tasks.

\subsection{Experiment 7: Extension to transformers}
\label{sec:exp7}

Experiments 1--6 establish gradient transport redundancy in recurrent architectures (vanilla RNNs and LSTMs), where the Jacobian propagation term $W_{hh} J_{t-1}$ is the mechanism under study. Does this extend to transformers? In transformers \citep{vaswani2017attention}, the structural analogue is \emph{sparse gradient transport}: restricting gradient flow through a subset of attention heads during backpropagation while maintaining the full forward pass.

\paragraph{Mechanism.} In standard multi-head attention \citep{vaswani2017attention}, the output is $y = W_O \, [\text{head}_1; \ldots; \text{head}_H]$. In sparse gradient transport, all $H$ heads compute their attention and contribute to the output projection normally. During backpropagation, non-selected heads are \emph{detached} from the computation graph:
\begin{equation}
    y = W_O \, [\tilde{h}_1; \ldots; \tilde{h}_H], \quad \tilde{h}_i = \begin{cases} \text{head}_i & \text{if } i \in S \\ \text{sg}(\text{head}_i) & \text{otherwise} \end{cases}
    \label{eq:sparse_head}
\end{equation}
where $S$ is a random subset of $k$ heads (resampled each forward pass) and $\text{sg}(\cdot)$ denotes stop-gradient. Note that $S$ is resampled stochastically, unlike the RNN experiments where the mask $M$ is fixed at initialization. This stochastic selection is closer to dropout-style regularization and may contribute to the observed generalization benefit at $k{=}6$. The forward computation uses all heads; only gradient transport is restricted to $k$ paths.

\paragraph{Setup.} We use a vision transformer \citep{dosovitskiy2021vit} on Split CIFAR-10 (5 sequential tasks of 2 classes each), a standard continual learning benchmark \citep{zenke2017continual}. The model uses $H{=}12$ heads (embed\_dim$=$192, head\_dim$=$16, depth$=$4) with a class-balanced replay buffer (1000 samples, $\lambda{=}5.0$). Sparse gradient transport is applied to all transformer blocks. We sweep $k \in \{1, 2, 4, 6\}$ across 3 seeds, with recovery computed as $(\text{Acc}_{\text{sparse}} - \text{Acc}_{\text{degenerate}}) / (\text{Acc}_{\text{dense}} - \text{Acc}_{\text{degenerate}})$.

\begin{table}[h]
\centering
\caption{Sparse gradient transport on Split CIFAR-10 ($H{=}12$ heads, 3 seeds). All heads contribute to the forward pass; gradient flows through only $k$ heads. At $k{=}6$ (50\%), sparse transport \emph{outperforms} the dense reference (regularization effect). The threshold for $>$70\% recovery is ${\sim}33\%$ of heads, higher than RNNs (${\sim}6\%$), reflecting greater head specialization.}
\label{tab:transformer}
\begin{tabular}{lcccccc}
\toprule
 & & \multicolumn{3}{c}{Avg Accuracy} & & \\
\cmidrule(lr){3-5}
Method & $k/H$ & Seed 42 & Seed 43 & Seed 44 & Mean & Recovery \\
\midrule
bptt\_ref & 12/12 & 0.407 & 0.397 & 0.374 & 0.393 & --- \\
dense\_ref & 12/12 & 0.403 & 0.389 & 0.401 & 0.398 & 100\% \\
sparse $k{=}6$ & 6/12 & 0.406 & 0.382 & 0.415 & 0.401 & 107\% \\
sparse $k{=}4$ & 4/12 & 0.387 & 0.376 & 0.384 & 0.383 & 77\% \\
sparse $k{=}2$ & 2/12 & 0.353 & 0.325 & 0.336 & 0.338 & 8\% \\
trace\_like & 1/12 & 0.328 & 0.317 & 0.346 & 0.330 & 0\% \\
\bottomrule
\end{tabular}
\end{table}

The results (Table~\ref{tab:transformer}) show gradient transport redundancy in transformers with a threshold between 17\% and 50\% of heads. At $k{=}6$ (50\%), sparse transport achieves 107\% recovery, \emph{outperforming} the dense reference. At $k{=}4$ (33\%), recovery is 77\% across 3 seeds (per-seed: 80\%, 82\%, 69\%), but 2 additional seeds show compressed or inverted dense-trace gaps (dense $\leq$ trace), so recovery metrics there are unreliable. $k{=}4$ is genuinely borderline. Below the threshold, performance degrades rapidly: $k{=}2$ (17\%) achieves only 8\% recovery.

The mechanism differs from RNNs. In RNNs, the Jacobian is near-isotropic and any random subset is representative (Section~\ref{sec:discussion}). In transformers, attention heads are functionally specialized \citep{voita2019heads}, where each learns distinct attention patterns, so gradient signal is concentrated in a subset of heads. Redundancy arises not from isotropy but from the fact that a random sample of 4--6 of 12 heads is likely to include enough of the important ones. This predicts (and Experiment~8 confirms) that head selection \emph{should} matter in transformers, unlike in RNNs.

\paragraph{Comparison with sparse forward transport.} We also test a variant where non-selected heads are \emph{zeroed} in the forward pass rather than detached, affecting both forward computation and gradient flow (analogous to structured head dropout). This yields a higher apparent threshold of ${\sim}50\%$ for $>$70\% recovery. The difference quantifies forward-pass corruption: with zeroing, the degenerate baseline drops much further (accuracy 0.14 vs 0.33 with detach), creating a larger gap but one that conflates feature corruption with gradient information. The 17 percentage point threshold reduction (50\% $\to$ 33\%) from isolating the gradient-only effect shows that the true gradient redundancy threshold is lower than naive forward sparsification suggests.

\paragraph{Forgetting reduction.} Sparse gradient transport substantially reduces catastrophic forgetting. At $k{=}6$, average forgetting is 0.22 versus 0.45 for the dense reference, a ${\sim}50\%$ reduction. This regularization effect is consistent across sparsity levels and across both sparse gradient and sparse forward transport variants.

\subsection{Experiment 8: Head selection matters in transformers}
\label{sec:exp8}

In RNNs, the choice of which $k$ Jacobian paths to propagate does not matter (Section~\ref{sec:exp3}): oracle, random, and adversarial selection all achieve equivalent recovery. Does the same hold for transformer heads?

We test this using Fisher information to identify the most and least important attention heads. After training on task 1 (the first 2-class split), we compute per-head Fisher importance by decomposing the squared gradient of the classification loss across heads in each layer's $W_Q$, $W_K$, $W_V$, and $W_O$ parameters. We then compare three selection strategies for tasks 2--5, all at $k{=}6$ of $H{=}12$: oracle (top-6 by Fisher), random (6 heads uniformly at random, resampled each forward pass), and anti-oracle (bottom-6 by Fisher, deliberately the 6 \emph{least} important heads).

\begin{table}[h]
\centering
\caption{Head selection strategies on Split CIFAR-10 ($H{=}12$, $k{=}6$, 5 seeds). Fisher oracle selects heads with the highest squared-gradient importance; entropy oracle selects heads with the lowest attention entropy (most peaked attention, i.e., least dormant). The Fisher oracle--anti gap of $+$24.3pp contrasts sharply with RNNs ($-0.6$pp; see text), while entropy oracle $\approx$ random shows dormancy-based selection adds no useful signal. Seed 2024 is an outlier (Fisher anti recovery 88.3\%); the remaining 4 seeds show gaps of $+$25--34pp.}
\label{tab:fisher}
\begin{tabular}{lcccccc}
\toprule
 & \multicolumn{5}{c}{Recovery (\%)} & \\
\cmidrule(lr){2-6}
Method & Seed 7 & Seed 42 & Seed 43 & Seed 44 & Seed 2024 & Mean \\
\midrule
Fisher oracle (top-6) & 96.9 & 97.6 & 93.7 & 89.2 & 99.1 & 95.3 \\
Entropy oracle (top-6) & 58.0 & 70.0 & 76.9 & 87.8 & 89.5 & 76.4 \\
Random & 84.0 & 84.2 & 66.3 & 80.6 & 72.5 & 77.5 \\
Entropy anti (bottom-6) & 82.3 & 61.1 & 82.8 & 66.9 & 85.3 & 75.7 \\
Fisher anti (bottom-6) & 70.1 & 63.3 & 69.7 & 63.7 & 88.3 & 71.0 \\
\bottomrule
\end{tabular}
\end{table}

The result is consistent across seeds: head selection matters in transformers, and the signal is in gradient importance, not activation patterns (Table~\ref{tab:fisher}). Fisher oracle selection achieves 95.3\% mean recovery while Fisher anti-oracle achieves only 71.0\%, a gap of $+$24.3pp that is positive on all 5 seeds ($+$26.8, $+$34.3, $+$24.0, $+$25.5, $+$10.8pp). This is the complete opposite of the RNN finding: a Fisher-based oracle experiment in RNNs (3 seeds) yields a gap of only $-0.6$pp (Fisher oracle 78.8\% vs.\ anti 79.4\%), with both underperforming random selection (88.7\%). Fisher importance varies up to 1000$\times$ across neurons yet carries no useful selection signal, which confirms Jacobian isotropy.

Entropy-based selection, which ranks heads by how peaked their attention is (effectively selecting the most ``active'' heads), performs no better than random (76.4\% vs.\ 77.5\%). This rules out a dormancy-based explanation of the Fisher gap. If Fisher oracle were merely selecting active heads over dormant ones, entropy oracle would show a comparable advantage; it does not. The Fisher--entropy oracle gap of $+$18.9pp (mean across 5 seeds; per-seed range $+$1.4 to $+$38.9pp) confirms that Fisher importance captures genuine gradient specialization (which heads carry disproportionate gradient signal for adaptation) beyond what activation-based measures detect. Head dormancy is transient: by later tasks, nearly all selected heads are active. The dormancy observed at initialization resolves during training.

This confirms the prediction from Section~\ref{sec:exp7}: because transformer heads are specialized rather than isotropic, selection matters. The opposite of the RNN finding ($-0.6$pp Fisher gap, Section~\ref{sec:exp3}).

This finding is distinct from the head pruning literature \citep{voita2019heads, michel2019heads}, which asks which heads can be removed \emph{post-training} without degrading accuracy. We ask a different question: which heads carry useful gradient signal \emph{during training} for adaptation to distribution shifts. The Fisher oracle--anti gap shows that gradient importance computed after task~1 predicts gradient utility on subsequent tasks. This connects head specialization to gradient transport redundancy across task boundaries.

\paragraph{Fisher timing sensitivity.} We test the robustness of Fisher selection by expanding to 5 seeds (42--46) and comparing Fisher computed at different task boundaries. Fisher importance computed after task~1 yields $81.6 \pm 4.0\%$ recovery on tasks 3--5 ($+$22.5pp oracle--anti gap), but Fisher computed after task~2 degrades to $45.5 \pm 7.1\%$ on the same tasks, \emph{worse} than random selection ($50.9\%$). Early Fisher (after task~1) captures general head importance before task-specific specialization sets in; later Fisher overfits to the most recent distribution. Head importance rankings are statistically correlated across tasks (Spearman $\rho = 0.78$), but small rank shifts near the selection threshold change which heads are chosen. This timing sensitivity is a limitation of Fisher-guided selection, not of the underlying redundancy finding; random selection remains robust regardless of timing.

\subsection{Experiment 9: Cross-session neural decoding}
\label{sec:exp9}

Experiments 1--8 establish gradient transport redundancy on synthetic dynamical tasks and image classification benchmarks. Does the finding transfer to real-world distribution shifts? We test on brain-computer interface (BCI) neural decoding, where electrode drift between recording sessions degrades decoder performance, a clinically significant problem that currently requires manual recalibration.

\paragraph{Setup.} We use the Indy reaching dataset \citep{odoherty2017reaching, makin2018bci}: 96-channel Utah array recordings from a macaque performing a center-out reaching task. We select two sessions separated by 7 months (session A: June 2016, session B: January 2017) to maximize electrode drift. Neural activity is reduced to 40 PCA dimensions (fit on session A only) and z-scored using session A statistics; session B data is processed with session A's transform, so any distributional shift from electrode drift is preserved. A vanilla RNN decoder ($n{=}64$, same architecture and optimizer as Experiments 1--6) is trained on session A using each gradient method, then evaluated in a single online pass through session B with no recalibration. The frozen baseline uses the session-A-trained model without adaptation.

\paragraph{Recovery metric.} For BCI, we define recovery relative to the frozen decoder's degradation rather than using Eq.~\ref{eq:recovery}:
\begin{equation}
    \text{Recovery}_{\text{BCI}} = \frac{\text{MSE}_{\text{frozen}}^B - \text{MSE}_{\text{method}}^B}{\text{MSE}_{\text{frozen}}^B - \text{MSE}_{\text{frozen}}^A} \times 100\%
    \label{eq:bci_recovery}
\end{equation}
where $\text{MSE}_{\text{frozen}}^A$ is the frozen model's end-of-session-A performance (the target) and $\text{MSE}_{\text{frozen}}^B$ is its degraded session-B performance (the floor). This measures what fraction of the drift-induced degradation each method recovers through online adaptation. Note that this metric is not directly comparable to the log-scale gap recovery in Eq.~\ref{eq:recovery}: ``80\% recovery'' on BCI and ``84\% recovery'' on sine measure different quantities.

\begin{table}[h]
\centering
\caption{Cross-session neural decoding (7-month electrode drift, $n{=}64$, 5 seeds). Frozen decoder degrades massively ($+$0.29--0.47 MSE). Sparse RTRL ($k{=}4$) recovers $80 \pm 11\%$ of the drift gap, with lower variance than full RTRL. Full RTRL is unstable on 2 of 5 seeds (bold), reproducing the pattern from chaotic dynamics (Table~\ref{tab:lorenz}).}
\label{tab:bci}
\begin{tabular}{lccccc}
\toprule
 & \multicolumn{5}{c}{Late session B MSE} \\
\cmidrule(lr){2-6}
Method & Seed 42 & Seed 123 & Seed 31337 & Seed 7 & Seed 2024 \\
\midrule
Frozen (no adapt.) & 0.734 & 0.547 & 0.674 & 0.628 & 0.580 \\
$k{=}0$ (traces) & 0.479 & 0.360 & 0.432 & 0.464 & 0.423 \\
$k{=}4$ (6\%) & 0.400 & 0.328 & 0.386 & 0.306 & 0.360 \\
$k{=}64$ (full RTRL) & 0.303 & \textbf{0.406} & 0.390 & 0.347 & \textbf{0.466} \\
BPTT (ref) & 0.351 & 0.329 & 0.330 & 0.317 & 0.333 \\
\midrule
$k{=}4$ recovery & 71.1\% & 75.6\% & 85.2\% & 97.5\% & 72.4\% \\
\bottomrule
\end{tabular}
\end{table}

Sparse RTRL adapts to real cross-session electrode drift (Table~\ref{tab:bci}). With $k{=}4$ (6\% of neurons), recovery is $80 \pm 11\%$ across 5 seeds, all seeds above the 60\% threshold. The frozen decoder degrades by $+$0.29 to $+$0.47 MSE, confirming substantial electrode drift. Sparse RTRL closes most of this gap through purely online adaptation, with no recalibration step and no stored session-B data.

The stability advantage of sparse over full RTRL reappears on real neural data. Full RTRL fails on 2 of 5 seeds (recovery 48.9\% and 37.6\%), while $k{=}4$ is stable on all 5 (recovery 71--98\%). The variance pattern matches the Lorenz finding (Section~\ref{sec:exp2}): $k{=}4$ Late-B MSE standard deviation is 0.037 versus 0.060 for full RTRL. On real biological signals, as on chaotic dynamics, the exact Jacobian is less reliable than a sparse approximation.

Eligibility traces ($k{=}0$) partially adapt on BCI data (58.5\% mean recovery), unlike on synthetic tasks where traces fail entirely. BCI electrode drift is slower and smoother than the abrupt frequency shifts in Experiments 1--2, allowing even crude one-step gradient estimates to track some of the distributional change. Still, $k{=}4$ outperforms traces by $+$22pp, and Jacobian propagation still provides substantial value even when the drift is gradual.

\begin{figure}[t]
\centering
\includegraphics[width=\textwidth]{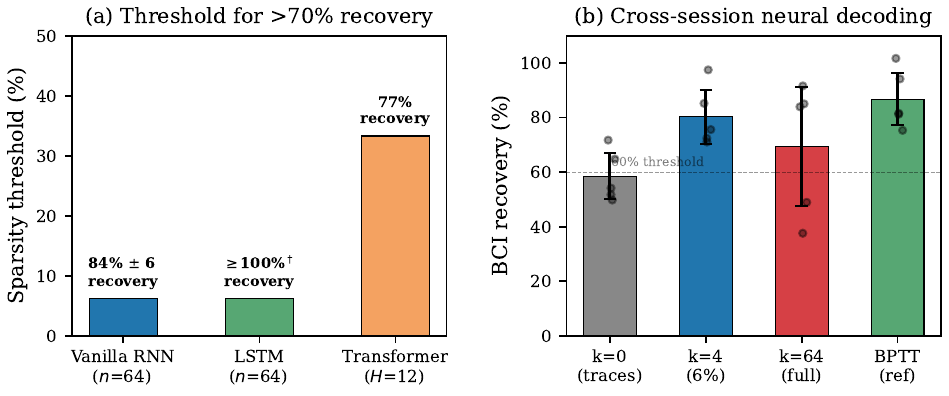}
\caption{Gradient transport redundancy across architectures and real neural data. (a) Recovery threshold comparison: RNNs require ${\sim}6\%$ of neurons, transformers ${\sim}33$--$50\%$ of heads. The difference reflects isotropy vs.\ specialization. (b) BCI cross-session adaptation: $k{=}4$ sparse RTRL adapts to 7-month electrode drift while the frozen decoder degrades and full RTRL is unstable on 2/5 seeds.}
\label{fig:cross_arch_bci}
\end{figure}

\section{Discussion}
\label{sec:discussion}

\subsection{Why is the Jacobian redundant?}

The central empirical finding of this paper, that random sparse Jacobian propagation recovers nearly all of full RTRL's adaptation ability, raises the question: why? A natural hypothesis is that the Jacobian tensor is low-rank along the neuron axis: if the effective rank is $r \ll n$, then any $k \geq r$ random samples would span the relevant subspace, and that would explain why $k{=}4$ suffices.

We tested this by computing the singular value decomposition of the Jacobian tensor $J_t$, reshaped from $(n, n, n)$ to $(n, n^2)$, at regular intervals during training. The low-rank hypothesis fails: at $n{=}64$, the effective rank at the 95\% variance threshold is $r_{95} \approx 60$; the Jacobian is essentially full rank. At $n{=}128$, $r_{95} \approx 119$. The Jacobian does not concentrate sensitivity in a low-dimensional subspace.

The spectral analysis does reveal a different and more explanatory property: the Jacobian is \emph{near-isotropic}. The condition number (ratio of largest to smallest singular value) is small, between 2.6 and 6.5 across all tasks and seeds (Table~\ref{tab:spectral}). The singular values are nearly uniform, meaning all neurons carry sensitivity information of similar magnitude. There are no privileged directions. This isotropy is not an initialization artifact: tracking condition numbers through training on the sine task ($n{=}64$, 5 seeds) shows they remain in the range 1.9--5.0 from $t{=}500$ through $t{=}1950$, increasing modestly in late training (mean 2.9 at the shift point, 3.9 at $t{=}1950$) as the network develops some directional structure, but never approaching the regime ($>$10) where random subsampling would degrade. Effective rank ($r_{95} \approx 60/64$) is stable throughout.

\begin{table}[h]
\centering
\caption{Spectral analysis of the Jacobian tensor along the neuron axis. The Jacobian is full-rank but near-isotropic: condition numbers of 2.6--6.5 indicate nearly uniform singular values. Values are mean $\pm$ s.d.\ where multiple seeds are available.}
\label{tab:spectral}
\begin{tabular}{lccccc}
\toprule
Task & $n$ & Seeds & $r_{95}$ & Cond.\ ($\sigma_1/\sigma_n$) \\
\midrule
Sine & 64 & 5 & $59.8 \pm 0.4$ & $3.97 \pm 0.81$ \\
Lorenz & 64 & 3$^\dagger$ & $56.3 \pm 0.5$ & $2.95 \pm 0.36$ \\
Sine & 128 & 5 & $117.6 \pm 0.5$ & $6.51 \pm 0.26$ \\
Sine & 256 & 3 & $236.7 \pm 0.4$ & $2.97 \pm 0.33$ \\
\bottomrule
\multicolumn{5}{l}{\footnotesize $^\dagger$3 of 5 seeds; 2 seeds diverged (condition number $>10^4$; see text).} \\
\end{tabular}
\end{table}

\begin{figure}[t]
\centering
\includegraphics[width=\textwidth]{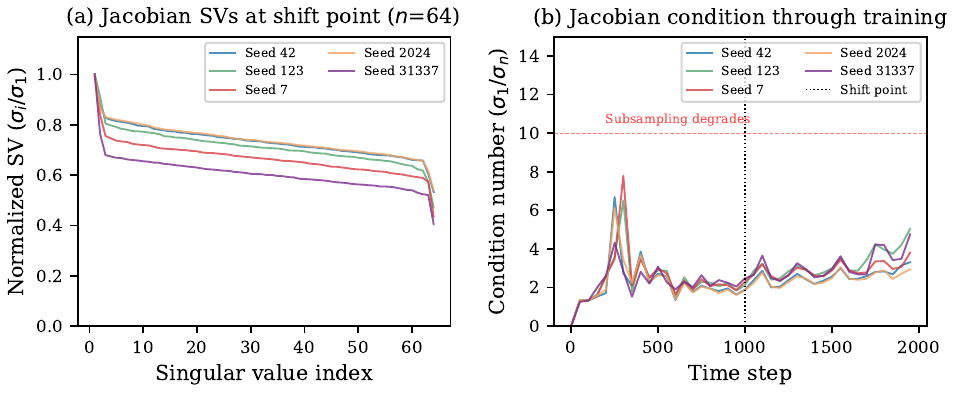}
\caption{The Jacobian is full-rank and near-isotropic. (a) Singular value spectrum of the Jacobian tensor (reshaped to $n \times n^2$) at the shift point ($n{=}64$, sine task), with nearly uniform singular values (condition number ${\sim}4$). Individual seeds shown. (b) Condition number remains low (2--5) throughout training across all 5 seeds. Isotropy is not an initialization artifact. Red dashed line: condition $= 10$, above which random subsampling would degrade.}
\label{fig:spectral}
\end{figure}

The 5-seed spectral data also reveals why full RTRL diverges on certain Lorenz seeds: the 2 seeds where full RTRL failed (Table~\ref{tab:lorenz}, seeds 7 and 31337 at $n{=}64$) are exactly the seeds where the Jacobian spectrum collapses, with condition numbers exceeding $10^4$. On these seeds, the Jacobian loses its isotropy and concentrates sensitivity into a few dominant singular values. Full RTRL faithfully propagates this pathological spectrum; sparse propagation, by subsampling, avoids amplifying the dominant modes.

When the Jacobian is isotropic, the experimental results follow directly. At $k{=}0$, there is no recurrent sensitivity information at all, so adaptation fails. At $k{=}4$, any random sample from a near-isotropic distribution preserves gradient direction. Magnitudes at $k{=}4$ are comparable to full RTRL (median ratio ${\approx}1.0$), so magnitude compensation is unnecessary; an SGD ablation confirms this with $92 \pm 1\%$ recovery across 5 seeds (see Limitations). From $k{=}4$ to $k{=}64$, additional samples add magnitude but no new directional information, which explains the flat recovery curve. And because all singular values are similar, no subset is privileged, which is why selection does not matter.

Direct gradient measurement confirms this: cosine similarity between $k{=}4$ and full RTRL gradients is $0.92 \pm 0.02$ pre-shift and $0.82 \pm 0.02$ post-shift (5 seeds), with a magnitude ratio of ${\sim}0.84$. Even during rapid adaptation after a distribution shift, sparse gradients preserve ${>}80\%$ directional alignment with the exact gradient. Cosine similarity increases monotonically with $k$ ($0.82 \to 0.85 \to 0.89$ for $k{=}4, 8, 16$ post-shift), consistent with the flat recovery curve: most directional information is captured by the first few paths.

\begin{figure}[t]
\centering
\includegraphics[width=\textwidth]{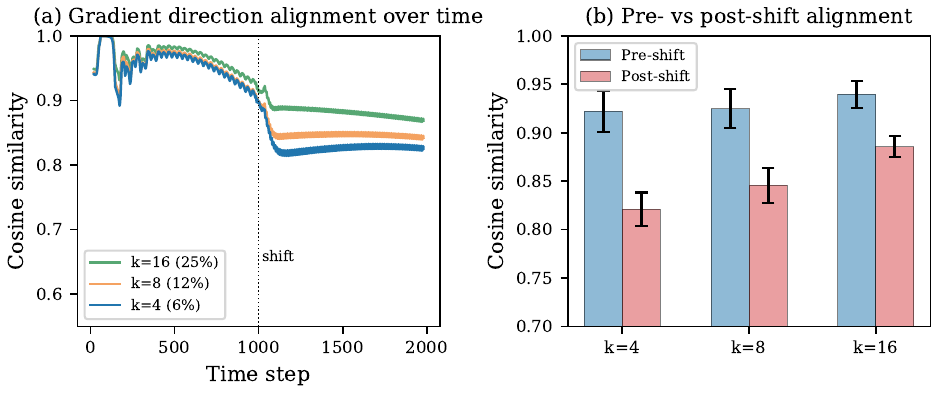}
\caption{Sparse gradients are directionally aligned with full RTRL gradients. (a) Cosine similarity between $k$-sparse and full RTRL gradients over time (5-seed mean $\pm$ s.d.). Pre-shift cosine is ${\sim}0.92$; post-shift dip to ${\sim}0.82$ at $k{=}4$ reflects the harder adaptation regime but never collapses. (b) Post-shift cosine increases monotonically with $k$, consistent with the flat recovery curve.}
\label{fig:cosine}
\end{figure}

What property of gradient-based training produces and maintains this near-isotropy remains an open question. Dynamical isometry theory shows that orthogonal initialization can produce near-uniform singular values at $t{=}0$, and edge-of-chaos results show that training drives the spectral radius toward 1, but neither explains why the \emph{full} spectrum remains near-uniform throughout training rather than developing large outlier singular values. Tanh saturation may act as a soft equalizer (neurons driven toward saturation by large singular value modes have derivatives near zero, selectively dampening the dominant modes), but this has not been formalized.

This connects to random projection theory \citep{johnson1984jl}: random projections preserve distance structure most reliably when the source distribution is isotropic, because the projection error is uniform across all directions. The near-isotropy of the Jacobian is precisely the condition under which random subsampling is most robust. The condition numbers on Lorenz ($2.95 \pm 0.36$ at $n{=}64$) are actually lower than on sine ($3.97 \pm 0.81$), so the more variable recovery on chaotic dynamics reflects numerical instability of the Jacobian \emph{magnitude} rather than degraded isotropy.

\subsection{Architecture-dependent redundancy thresholds}

The cross-architecture comparison (Table~\ref{tab:cross_arch}) shows that gradient transport redundancy is a general phenomenon, but the degree of redundancy varies by architecture.

\begin{table}[h]
\centering
\caption{Gradient transport redundancy across architectures. The threshold for $>$70\% recovery varies from ${\sim}6\%$ (RNNs) to ${\sim}33\%$ (transformers), reflecting different specialization structures. Transformer threshold measured with sparse gradient transport (Section~\ref{sec:exp7}).}
\label{tab:cross_arch}
\begin{tabular}{lcccc}
\toprule
Architecture & Transport unit & Threshold & Recovery at threshold & Degenerate \\
\midrule
Vanilla RNN & Neuron & $k{=}4$/64 (6\%) & $84 \pm 6\%$ & $k{=}0$ (fails) \\
LSTM & Gate Jacobian & $k{=}4$/64 (6\%) & $\geq$100\% & $k{=}0$ (fails) \\
Transformer & Attention head & $k{=}4$--6/12 (33--50\%) & 77--107\%$^\dagger$ & $k{=}1$ (0\%) \\
\bottomrule
\multicolumn{5}{l}{\footnotesize $^\dagger$$k{=}4$ is borderline (high seed variance); $k{=}6$ is robust (107\% recovery, 3 seeds).}
\end{tabular}
\end{table}

In RNNs, the near-isotropic Jacobian (condition numbers 2.6--6.5) means all neurons carry similar sensitivity information. Any small random subset provides a representative gradient estimate. The threshold is correspondingly low: 4 of 64 neurons, regardless of which 4 (Section~\ref{sec:exp3}).

In transformers, the mechanism is different: heads are functionally specialized (Section~\ref{sec:exp7}), so gradient signal is concentrated rather than uniformly distributed. The Fisher head selection experiment (Section~\ref{sec:exp8}) confirms this with a $+$24.3pp oracle--anti gap across 5 seeds (vs.\ $-0.6$pp in RNNs), while entropy-based selection matches random. The threshold is correspondingly higher: ${\sim}33\%$ of heads for gradient-only sparsity, or ${\sim}50\%$ when forward-pass corruption is also present. The threshold difference is confounded with the number of transport units: in absolute terms, both architectures require $k \approx 4$. Disentangling the effects of unit count and unit specialization would require testing transformers with substantially more heads (e.g., 64) at fixed $k$.

We emphasize that these are analogous phenomena under a shared framing (gradient transport redundancy), not the same mechanism in different architectures: the RNN result concerns forward-mode Jacobian propagation redundancy, while the transformer result concerns backward-mode gradient transport through attention heads. Despite these quantitative and mechanistic differences, the qualitative pattern is the same across all three architectures: (1) a degenerate baseline where gradient transport is eliminated or near-eliminated collapses completely, (2) there exists a threshold above which recovery is substantial ($>$70\%), and (3) below this threshold, performance degrades rapidly. The existence of a redundancy threshold holds across all three architecture families tested; the threshold level is architecture-specific and predictable from the specialization structure of the transport units.

\subsection{Implicit regularization on chaotic dynamics}

On the Lorenz attractor (Section~\ref{sec:exp2}), sparse Jacobian propagation is \emph{more numerically stable} than full RTRL. With 5 seeds, the pattern is clear: $k{=}4$ achieves consistent post-shift MSE ($0.015 \pm 0.002$ at $n{=}64$, CV $= 13\%$) while full RTRL varies wildly ($0.036 \pm 0.035$, CV $= 88\%$), diverging entirely on 2 of 5 seeds. At $n{=}128$, the instability worsens: full RTRL's CV reaches 161\% and it diverges on 3 of 5 seeds, while $k{=}4$ remains tight ($0.016 \pm 0.001$, CV $= 8\%$).

In a chaotic system, nearby trajectories diverge exponentially. \citet{mikhaeil2022chaos} prove that loss gradients of RNNs producing chaotic dynamics always diverge, regardless of architecture. The RTRL Jacobian tensor inherits this instability: the sensitivity $\partial h_t / \partial \theta$ grows exponentially with $t$, even when the gradient signal it encodes is bounded. Full RTRL propagates this exponential growth faithfully, injecting multiplicative noise into the gradient. Sparse propagation, by summing over only $k$ of $n$ terms per row, implicitly dampens this growth, a form of gradient regularization. The spectral analysis (Table~\ref{tab:spectral}) confirms the mechanism: on the seeds where full RTRL diverges, the Jacobian's condition number exceeds $10^4$, indicating catastrophic loss of isotropy. Sparse subsampling avoids amplifying these pathological modes.

This observation connects to a broader point: the ``exact'' gradient is not always the best gradient. On chaotic dynamics, the exact RTRL gradient is numerically unstable, and approximations that introduce bias can outperform exact methods by reducing variance. UORO's rank-1 compression achieves an even stronger regularization effect ($0.0025 \pm 0.0006$ on Lorenz, outperforming all other methods), which explains its advantage on chaotic tasks, at the cost of destroying useful information on non-chaotic tasks ($0.049$ on multi-sine vs $0.005$ for $k{=}4$).

\subsection{When forward-mode breaks down}

Experiment~4 (Section~\ref{sec:exp4}) reveals a scope condition for all RTRL-based methods: without continuous error signal, the Jacobian accumulates numerical drift during ungrounded propagation, and full RTRL becomes the \emph{worst} method, underperforming even eligibility traces. This predicts that RTRL-based methods will degrade on any task with temporally sparse error signals (reinforcement learning with delayed rewards, sequence-to-sequence with decoder-only loss). The Jacobian needs continuous grounding to remain useful. No prior RTRL work has explicitly characterized this scope condition, and it constrains the applicability of the entire family of forward-mode approaches.

\subsection{Implications for RTRL approximation design}

Our results suggest a reframing of the design space for RTRL approximations. Prior work has invested significant engineering effort in designing \emph{structured} approximations: graph-based sparsity patterns that follow network topology (SnAp), optimal matrix factorizations that minimize approximation error (OK), architectural constraints that simplify the Jacobian structure (Irie). Each of these methods implicitly assumes that the \emph{choice} of which Jacobian information to retain matters.

Our selection invariance result (Section~\ref{sec:exp3}) challenges this assumption. The information in the recurrent Jacobian is so redundant that even adversarial selection, deliberately retaining only the weakest connections, works as well as oracle selection. This does not invalidate prior methods: SnAp, UORO, and OK may offer advantages on tasks and architectures we have not tested. But it does suggest that the engineering effort devoted to structured selection may yield diminishing returns. A practitioner choosing an RTRL approximation should focus on two questions: (1) does it propagate \emph{any} Jacobian information? and (2) can the architecture solve the task? The specific structure of the approximation appears secondary.

\subsection{Limitations}

Our findings have clear boundaries, and the gaps matter.

On the architecture side, we tested three families (vanilla RNNs, LSTMs, vision transformers), each with different redundancy thresholds. LSTM RTRL did not converge on chaotic dynamics, consistent with \citet{mikhaeil2022chaos}. GRUs remain untested but are structurally similar to LSTMs. State-space models and linear recurrent units have qualitatively different Jacobian structure and may not exhibit the same isotropy. Our transformer is small (4 blocks, 12 heads); we do not know whether the ${\sim}33\%$ threshold holds for large-scale models with 32+ heads.

On tasks, RNN experiments use continuous dynamical prediction with dense error signals; the transformer uses continual image classification with replay; the BCI experiment uses real neural decoding with cross-session drift. We have not tested language modeling or reinforcement learning. Experiment~4 shows that sparse error signals degrade all RTRL variants, and that boundary is sharp. The BCI experiment uses a single session pair from one subject; a clinical evaluation would need multiple session pairs, multiple subjects, and comparison against standard recalibration baselines.

On scale, the absolute-$k$ law holds from $n{=}64$ to $n{=}256$. At $n{=}512$, RTRL itself breaks: the full Jacobian diverges. Networks above $n{=}1000$, where the computational savings would matter most, likely need mixed-precision or periodic Jacobian renormalization. We see no way around this with standard floating-point arithmetic.

Seed counts vary: 5 seeds for most RNN experiments and Fisher selection, 3 seeds for the main transformer results (2 additional seeds showed compressed dense-trace gaps, confirming $k{=}4$ is borderline). The Lorenz spectral analysis excludes 2 of 5 diverged seeds; the $n{=}256$ spectral analysis uses 3 seeds.

Our evaluation protocol (online adaptation to distribution shifts) directly tests temporal credit assignment but lacks established baselines in the RTRL literature, which makes direct comparison with published results difficult. We include standard benchmarks in Experiment~4, where the comparison points to the architecture rather than the gradient method as the bottleneck.

All main RNN experiments use Adam. An SGD ablation on sine ($n{=}64$, 5 seeds) gives $92 \pm 1\%$ recovery (per-seed: 91.0\%, 93.5\%, 93.1\%, 91.3\%, 90.8\%), with low variance. Gradient magnitude diagnostics show the median $k{=}4$/$k{=}64$ ratio is ${\approx}1.0$ for both optimizers, so sparse gradients match full RTRL in magnitude and Adam's per-parameter normalization is not compensating for any reduction. But this is one task only; broader SGD validation is still needed.

\section{Related Work}
\label{sec:related}

\paragraph{Exact RTRL.} Real-time recurrent learning \citep{williams1989rtrl} computes exact forward-mode gradients by maintaining the full Jacobian tensor. Its $O(n^4)$ per-step cost has historically limited it to small networks, motivating the approximation methods below. \citet{irie2024elstm} recently achieved exact RTRL at $O(n^2)$ by eliminating inter-neuron recurrence entirely, a different tradeoff: architecture-level constraint rather than gradient-level approximation. Our LSTM results (Section~\ref{sec:exp5}) suggest that for standard architectures, sparse gate Jacobian propagation gets most of the benefit without changing the network.

\paragraph{Dropping the Jacobian.} RFLO \citep{murray2019rflo} and e-prop \citep{bellec2020eprop} drop the Jacobian propagation term entirely, retaining only one-step derivatives. This is our $k{=}0$ baseline (eligibility traces). These methods are biologically motivated and $O(n^2)$, but they cannot adapt to distribution shifts because they lose the recurrent gradient component that carries information across time steps.

\paragraph{Structured Jacobian approximations.} UORO \citep{tallec2017uoro} maintains a rank-1 approximation at $O(n^2)$ cost. On chaotic dynamics the high variance helps (Section~\ref{sec:exp2}); on non-chaotic tasks it destroys useful signal. \citet{benzing2019ok} derive an optimal Kronecker-sum approximation (OK) matching BPTT on Penn Treebank at $O(n^3)$. \citet{marschall2020unified} provide a unified taxonomy placing all of these on a spectrum from exact to maximally approximate.

\paragraph{SnAp: graph-structured sparsity.} SnAp \citep{menick2021snap} is the closest prior work: it also sparsifies the Jacobian propagation. The difference is that SnAp uses \emph{graph-structured} sparsity tied to the network's learned connectivity. For a weight-sparse RNN, each neuron propagates Jacobian information only through neurons reachable within $s$ hops in the weight graph. The sparsity pattern is deterministic and topology-dependent, and the network weights must themselves be sparse.

We keep $W_{hh}$ fully dense and sparsify the Jacobian propagation with a random subset that bears no relation to network topology. Our selection invariance result (Section~\ref{sec:exp3}) predicts that SnAp's structured selection may be unnecessary: random selection works equally well, and even adversarial selection cannot degrade performance. This is testable in SnAp's framework. SnAp evaluates on GRU architectures with binary copy and WikiText-2, so direct numerical comparison with our vanilla tanh RNNs on continuous tasks is not possible, but the mechanistic point stands: SnAp designs a specific structure; we show that structure is unnecessary.

\paragraph{Diagonal and linear recurrences.} \citet{zucchet2023online} combine linear recurrent units (LRUs) with RTRL, exploiting diagonal recurrence to reduce cost. This eliminates the inter-neuron Jacobian by design. \citet{wang2026braintrace} exploit biological activity sparsity in spiking neural networks: most neurons are silent at any given time, so the Jacobian is inherently sparse, enabling linear-memory online learning at whole-brain scale. Our networks have dense activations and a dense, full-rank Jacobian. The sparsity we exploit is not in the Jacobian itself but in how much of it needs to be \emph{propagated}.

\paragraph{Sparse gradient transport in transformers.} Our sparse gradient transport (Section~\ref{sec:exp7}) detaches non-selected attention heads during backpropagation while keeping the full forward pass. The question is different from dropout (which zeroes forward activations), DropHead \citep{zhou2020drophead} (structured dropout over heads), and head pruning \citep{voita2019heads, michel2019heads} (post-training removal of redundant heads). We are asking whether gradient transport through \emph{all} heads is necessary during training. Concurrently, \citet{guo2026greedygnorm} found that gradient-norm-based head scoring outperforms entropy for post-training pruning, paralleling our Experiment~8 result (Fisher $>$ entropy) in an independent setting. Gradient-based head importance may generalize across the training/inference divide. That ${\sim}33\%$ of heads suffice for $>$70\% recovery, combined with the Fisher result (Section~\ref{sec:exp8}), shows transformer heads are functionally specialized for gradient transport. The pruning literature has focused on inference-time redundancy; the training-time gradient information question has not been tested before.

\section{Conclusion}
\label{sec:conclusion}

Gradient transport in neural networks is massively redundant. In RNNs, a random 1.6--6\% of Jacobian paths recovers 78--84\% of full RTRL's adaptation, and the choice of which paths to keep does not matter: even adversarial selection works. The mechanism is near-isotropy. The Jacobian is full-rank but spectrally flat (condition numbers 2.6--6.5), so any random subset preserves gradient direction. This extends to LSTMs and, with higher thresholds reflecting head specialization rather than isotropy, to transformers ($+$24.3pp Fisher oracle--anti gap across 5 seeds, vs.\ $-0.6$pp in RNNs).

Two results were unexpected. First, on chaotic dynamics, sparse propagation is more stable than the exact method: $k{=}4$ holds a CV of 8--13\% while full RTRL diverges on 2--3 of 5 seeds. The approximation outperforms the thing it approximates, because subsampling dampens the exponential sensitivity growth that makes the exact Jacobian pathological. This same stability advantage reappears on real primate neural data (80\% recovery on 7-month electrode drift, all 5 seeds above threshold, full RTRL unstable on 2). Second, $k{=}4$ suffices regardless of network size, from $n{=}64$ (6\%) to $n{=}256$ (1.6\%). The savings grow with scale while recovery degrades modestly. At $n{=}256$, sparse RTRL propagates 64$\times$ less information while full RTRL shows 22$\times$ seed-dependent variance.

The price is a sharp scope condition. Without continuous error signal, the Jacobian accumulates drift and full RTRL becomes the \emph{worst} method, underperforming even eligibility traces. Forward-mode sensitivity methods need dense error feedback. This has not been characterized before, and it constrains the entire family of approaches, not just our sparse variant.

For practitioners designing RTRL approximations: the specific structure probably does not matter. Propagate any small random subset of the Jacobian, and the gradient will point approximately the right way. The question that matters is not which paths to keep, but whether you have any paths at all.

\paragraph{Future work.} Three directions are most pressing. First, scaling sparse gradient transport to transformers with 32+ heads would test whether the ${\sim}33\%$ threshold decreases with more heads (as it does for RNN neurons) or remains constant, disentangling the effects of unit count and specialization. Second, extending to language modeling and reinforcement learning would test whether gradient transport redundancy holds under different loss structures and sequence statistics. Third, a formal theoretical bound that connects Jacobian isotropy to gradient direction error (as a function of $k$, $n$, and the condition number) would place these empirical findings on rigorous theoretical ground. On the applied side, the BCI cross-session result (Section~\ref{sec:exp9}) motivates a full clinical evaluation: multiple session pairs at varying time gaps, multiple subjects, and comparison against standard recalibration methods, potentially enabling continuous decoder adaptation without manual recalibration.

\bibliography{references}
\bibliographystyle{plainnat}

\end{document}